\documentclass[runningheads]{llncs}
\pdfoutput=1

\usepackage{graphicx}

\usepackage{changes}

\usepackage{graphicx}
\usepackage{bm}
\usepackage[utf8]{inputenc}		
\usepackage[english]{babel}		
\usepackage{xspace}
\usepackage{amsmath}					
\usepackage{algorithm}
\usepackage{multirow}					
\usepackage{colortbl}					
\usepackage{tabularx}
\usepackage[noend]{algpseudocode}	
\usepackage{stackrel}

\usepackage{tikz}
\usetikzlibrary{shapes}
\usetikzlibrary{arrows}				
\usetikzlibrary{positioning}	
\usetikzlibrary{calc}					
\usetikzlibrary{matrix} 	    

\newcommand{\latexOrPdflatex}[2]{\ifx\undefined\pdfoutput%
#1%
\else%
#2%
\fi}

\ifdefined\printmode

\usepackage{url}

\else
\latexOrPdflatex{
\newcommand{\href}[2]{#2}
}{
\usepackage[colorlinks,citecolor=blue,bookmarks=true,pdfstartview=FitH,pdfstartpage=1,pdftitle={Final Adaptation Reinforcement Learning},pdfauthor={Wolfgang Konen and Samineh Bagheri}]{hyperref}     
}
\newcommand{\citep}[1]{\cite{#1}}

\renewcommand{\vec}[1]{\bm{#1}}	
\renewcommand{\hat}[1]{#1}	

\newcommand{\AfterstateLogic}[1]{{afterstate logic}\xspace}
\newcommand{\afterstate}[1]{{afterstate}\xspace}
\newcommand{\EligibilityMethod}[1]{{eligibility method}\xspace}
\newcommand{\NPlayerGames}[1]{{$N$-player games}\xspace}
\newcommand{\FinalAdaptStep}[1]{{final adaptation step}\xspace}

\newcommand{\Jaskowski}[1]{Ja{\'s}kowski\xspace}

\newcommand{\WK}[1]{\textcolor{blue}{(\textbf{WK:} \textit{#1})}}
\newcommand{\SB}[1]{\textcolor{orange}{(\textbf{SB:} \textit{#1})}}

\begin{document} 

\title{Final Adaptation Reinforcement Learning \\ for {N}-Player Games
\thanks{This work is an extended version of a paper submitted to \textit{BIOMA'2020, Bruxelles}.}
}
\titlerunning{Final Adaptation Reinforcement Learning}
%
\author{Wolfgang Konen
\orcidID{0000-0002-1343-4209} \and \\
Samineh Bagheri
\orcidID{0000-0003-2488-8000}
}
\authorrunning{W. Konen and S. Bagheri}
%
\institute{Computer Science Institute, TH Köln, University of Applied Sciences, Germany
\email{\{wolfgang.konen,samineh.bagheri\}@th-koeln.de}
}
\maketitle              
\begin{abstract}
This paper covers n-tuple-based reinforcement learning (RL) algorithms for games. We present new algorithms for TD-, SARSA- and Q-learning which work seamlessly on various games with arbitrary number of players. This is achieved by taking a player-centered view where each player propagates his/her rewards back to previous rounds. We add a new element called Final Adaptation RL (FARL) to all these algorithms. Our main contribution is that FARL is a vitally important ingredient to achieve success with the player-centered view in various games. We report results on seven board games with 1, 2 and 3 players, including Othello, ConnectFour and Hex. In most cases it is found that FARL is important to learn a near-perfect playing strategy.
All algorithms are available in the GBG framework on GitHub.
%

\end{abstract}

%


\section{Introduction} \label{sec:introduction}

\subsection{Motivation}

It is desirable to have a better understanding of the principles how computers can learn strategic decision making. Games are an interesting test bed and reinforcement learning (RL) is a general paradigm for strategic decision making.
It is however not easy to devise algorithms which work seamlessly on a large variety of games (different rules, goals and game boards, different number of players and so on). 
It is the hope that finding such algorithms and understanding which elements in them are important helps to better understand the principles of learning and strategic decision making. 

Learning how to play games with neural-network-based RL agents can be seen as a complex optimization task. It is the goal to find the right weights such that the neural network outputs the optimal policy for all possible game states or a near-optimal policy that minimizes the expected error. The state space in board games is usually discrete and in most cases too large to be searched exhaustively. These aspects pose challenges to the optimizer which has to generalize well to unseen states and has to avoid overfitting. 

In this paper we describe in detail a new class of n-tuple-based RL algorithms (TD-, SARSA- and Q-learning). N-tuples were introduced by Lucas \cite{Lucas08} to the field of game learning. These new learning algorithms extend the work described in~\cite{Bagh15,Kone15c,Thil14} and serve the purpose to be usable for a large variety of games. More specifically we deal here with discrete-time, discrete-action, one-player-at-a-time games. This includes board games and card games with $N=1,2,\ldots$ players. Games may be deterministic or nondeterministic. 


N-tuple networks are shown to work well in a variety of games, (e.g. in ConnectFour~\cite{Bagh15,Thil14}, Othello~\cite{Lucas08}, EinStein würfelt nicht~\cite{Chu2017EinStein}, 2048~\cite{szubert2014temporal}, SZ-Tetris~\cite{jaskowski2015high} etc.) but the algorithms described here are not tied to them. Any other function approximation network (deep neural network or other) could be used as well.

All algorithms presented here are implemented in the General Board Game (GBG) learning and playing framework~\cite{Konen2019b,Konen20a} and are applied to several games. The variety of games makes the RL algorithms a bit more complex than the basic RL algorithms. This paper describes the algorithm as simple as possible, yet as detailed as necessary to be precise and to follow the implementation in GBG's source code, which is available on GitHub\footnote{\url{https://github.com/WolfgangKonen/GBG}}.

A work related to GBG~\cite{Konen2019b,Konen20a} is the general game systems Ludii~\cite{Piette2019}. 
Ludii is an efficient general game system based on ludeme library implemented in Java, allowing to play as well as to generate a large variety of strategy games. Currently all AI agents implemented in Ludii are tree-based agents (MCTS variants or Alpha-Beta). GBG on the other hand offers the possibility to train RL-based algorithms on several games.

The main contributions of this paper are as follows: (i) It presents a unifying view for RL algorithms applicable to different games with different number of players; (ii) it demonstrates that a new element, named Final Adaptation RL (FARL), is vital for having success with this new unifying view;  (iii) it incorporates several other elements (afterstates, n-tuples, eligibility with horizon, temporal coherence) that are useful for all games. To the best of our knowledge, this is the first time that these elements are brought together in a comprehensive form for game-learning algorithms with arbitrary number $N$ of players. In addition, we point out and demonstrate  that SARSA- and Q-learning have some disadvantages in game learning as compared to TD-learning. 

\subsection{Algorithm Overview}
\label{sec:newLogic}

The most important task of a game-playing agent is to propose, given a game state $s_t$, a good next action $a_t$ from the set of available actions in $s_t$. 
TD-learning uses the value function $V(s_{t})$ which is the expected sum of future rewards when being in state $s_{t}$. Similarly, Q- and SARSA-learning (\textbf{S}tate-\textbf{A}ction-\textbf{R}eward-\textbf{S}tate-\textbf{A}ction) use the Q-function $Q(s_t,a_t)$ which is the expected sum of future rewards when taking action $a_t$ in state $s_t$. The similarities and differences between these variants are well explained in~\citep{Ree2013reinforcement,SuttBart98}.

It is the task of the agents to learn the value function $V(s)$ (TD) or  the Q-function $Q(s,a)$ (SARSA, Q) from experience (interaction with the environment). In order to do so, they usually perform multiple self-play training episodes, until a certain training budget is exhausted or a certain game-playing strength is reached. 

The nomenclature and algorithmic description follows as closely as possible the descriptions given in~\cite{jaskowski2018mastering,szubert2014temporal}. But these algorithms are for the special case of the 1-player game 2048. Since we want to use the TD-n-tuple algorithm for a broader class of games, we present in this paper a unified TD-update scheme inspired by~\cite{Ree2013reinforcement} which works for 1-, 2-, $\ldots$, $N$-player games.

Our new RL-algorithm is partly inspired by \cite{jaskowski2018mastering,Ree2013reinforcement} and partly from our own experience with RL\added{-}n-tuple training. The key elements of the new RL-logic -- as opposed to our previous RL-algorithms~\cite{Bagh15,Kone15c} -- are:
\begin{itemize}
	%
	\item New \AfterstateLogic\ \cite{jaskowski2018mastering}, see Sec.~\ref{sec:afterstate}.
	\item Generalization to \NPlayerGames\ with arbitrary $N$~\citep{Ree2013reinforcement}, see Sec.~\ref{sec:nPlayer}. 
	\item Final adaptation RL (FARL) for all players, see Sec.~\ref{sec:finalAdapt}. 
	\item Eligibility method with horizon based on~\cite{jaskowski2018mastering}, see Sec.~\ref{sec:elig-intro}.
\end{itemize}

The following key elements are also part of our algorithms, but only briefly explained in the appendix of this technical report

\begin{itemize}
	\item Weight-individual learning rates via temporal coherence learning (TCL)~\cite{Bagh15,Beal99}, see Appendix~\ref{sec:TCL}.
	\item Faster learning through symmetries~\cite{Lucas08}.
	\item Modified n-tuple update rule with global learning rate scaling factor. 
	see Appendix~\ref{sec:eligibility}.
	\item Inhibition of multiple changes of the same weight in one update step, see Appendix~\ref{sec:eligibility} and \ref{sec:helper}.
\end{itemize}


This technical report extends the BIOMA'2020 paper insofar that it covers SARSA as well, that it covers the aforementioned algorithmic details in Appendix~\ref{sec:eligibility}--\ref{sec:TCL}, and that it has in Appendix~\ref{sec:paramExp} all the parameter settings for the experiments shown in Tab.~\ref{tab:allGames}.

\section{Algorithms and Methods}
\label{sec:algoMeth}

\subsection{N-Tuple Systems}
N-tuple systems coupled with TD were first applied to game learning by Lucas in 2008~\cite{Lucas08}, although n-tuples were introduced already in 1959 for character recognition purposes. The remarkable success of n-tuples in learning to play Othello~\cite{Lucas08} motivated other authors to benefit from this approach for a number of other games. The main goal of n-tuple systems is to map a highly non-linear function in a low dimensional space to a high dimensional space where it is easier to separate 'good' and 'bad' regions. This can be compared to kernel trick in SVM. An n-tuple is defined as a sequence of $n$ cells of the board. Each cell can have $m$ values representing the possible states of that cell.  Therefore, every n-tuple will have a (possibly large) look-up table indexed in form of an $n$-digit number in base $m$. An n-tuple system contains multiple n-tuples.


\hypertarget{hrefAfterstate}{\subsection{Afterstate Logic}}
\label{sec:afterstate}

For nondeterministic games, Jaskowski et al.~\cite{jaskowski2018mastering,szubert2014temporal} describe a clever mechanism to reduce the complexity of the value function $V(s)$.

\begin{figure}%
\includegraphics[width=\columnwidth]{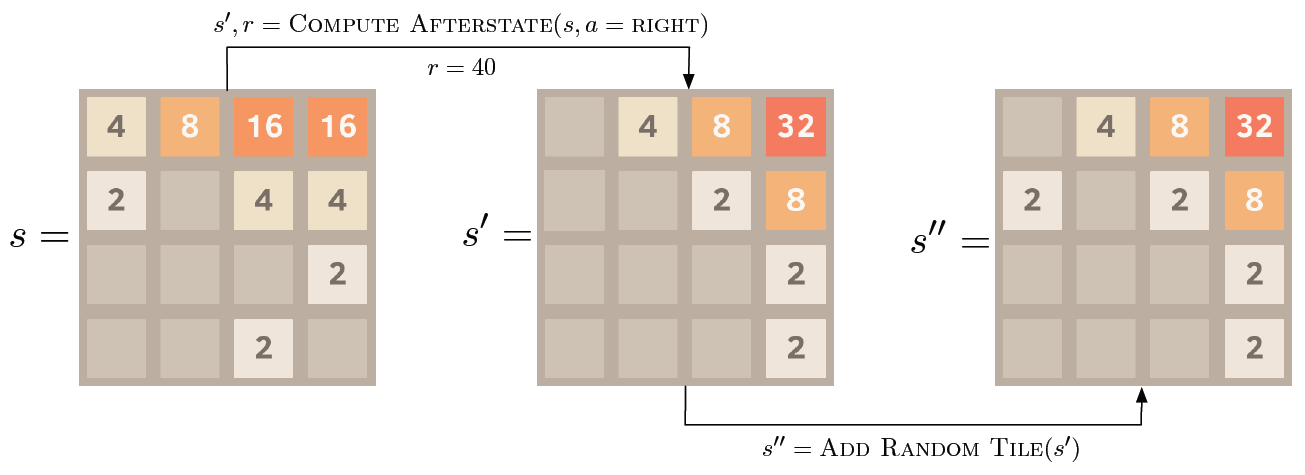}
\caption{For nondeterministic games it is better to split a state transition from $s$ to $s''$ in a deterministic part, resulting in \textbf{afterstate} $s'$, and a random part resulting in \textbf{next state} $s''$ (from~\cite{szubert2014temporal}).}%
\label{fig:afterstate2048}%
\end{figure}

Consider a game like 2048 (Fig.~\ref{fig:afterstate2048}): An exemplary action is to move all tiles to the right, this will cause the environment to merge adjacent same-value tiles into one single tile twice as big. This is the deterministic part of the action and the resulting state is called the \textbf{afterstate} $s'$. The second part of the action \textit{move-right} is that the environment adds a random tile 2 or 4 to one of the empty tiles. This results in the \textbf{next state} $s''$.

The naive approach for learning the value function would be to observe the next state $s''$ and learn $V(s'')$. But this has the burden of increased complexity: Given a state-action pair $(s,a)$ there is only \textit{one} afterstate, but $2n$ possible next states $s''$, where $n$ is the number of empty tiles in afterstate $s'$.\footnote{In the example of Fig.~\ref{fig:afterstate2048} we have $n=9$ empty tiles in afterstate $s'$, thus there are $2n=18$ possible next states $s''$. The factor 2 arises because the environment can place one of the two random tiles 2 or 4 in any empty tile.} This makes it much harder to learn the value of an action $a$ in state $s$. And indeed, it is not the specific value of $V(s'')$ which is the value of action $a$, but it is the expectation value 
$\left\langle V(s'') \right\rangle$ over \textit{all} possible next states $s''$.  

It is much more clever to learn the value $V(s')$ of an afterstate. This reduces the complexity by a factor of $2\bar{n}$, where $\bar{n}$ is the average number of empty tiles. It helps the agent to generalize better in all phases of TD-learning.  


For deterministic games there is no random part: afterstate $s'$ and next state $s''$ are the same. 
%
%
However, afterstates are also beneficial for deterministic games: For positional games (like TicTacToe, ConnectFour, Hex, $\ldots$) the value of taking action $a$ in state $s$ depends only on the resulting afterstate $s'$. Several state-action pairs might lead to the same afterstate, and it often reduces the complexity of game learning 
if we learn the the mapping from afterstates to game values (as we do in TD-learning, Sec.~\ref{sec:TDNTuple3Agt}). This is opposed to learning the mapping from state-action pairs to values (as we do in SARSA- and Q-learning, Sec.~\ref{sec:SarsaQLearn}). 



\subsection{Eligibility Method}
\label{sec:elig-intro}

Instead of Sutton's eligibility traces \citep{SuttBart98} we use in this work Jaskowski's eligibility method \cite{jaskowski2018mastering}. This method is efficiently computable even in the case of long RL episodes and it can be made equivalent to eligibility traces in the case of short episodes. For details the reader is referred to Appendix~\ref{sec:eligibility}  or to \cite{jaskowski2018mastering}. 

\hypertarget{hrefNPlayer}{\subsection{$N$ Players}}
\label{sec:nPlayer}


We want to propose a general TD($\lambda$) n-tuple algorithm which is applicable not only to 1- and 2-player games but to arbitrary 
$N$-player games. 

The key difference to the TD-learning variants described in earlier work~\citep{Bagh15,Kone15c} is that there each state was connected with the next state in the episode. This required different concepts for TD-learning, depending on whether we had a 1- or a 2-player game (minimization or maximization). Furthermore it has a severe problem for $N$-player games with $N>2$: We usually do not know  the game value for all other players in intermediate states, but we would need them for the algorithms in~\citep{Bagh15,Kone15c}. In contrast, Ree and Wiering~\cite{Ree2013reinforcement} describe an approach where each player has a value function only for \textit{his/her} states $s_t$ or state-action-pairs $(s_t,a_t)$. The actions of the opponents are subsumed in the reaction from the environment. That is, if $s_t$ is the state for player $p_t$ at time $t$, then $s_{t+1}$ is the next state of the \textit{same} player $p_t$ on which (s)he has to act. This has the great advantage that there is no need to translate the value of a state for player $p_t$ to the value for other players -- we take always the perspective of the same player when calculating temporal differences.\footnote{If the player uses a neural network for function approximation it may or may not be that the opponents also use the same network~\citep{Ree2013reinforcement}.}

This holds for TD-, SARSA- and Q-learning. 
In the next section we describe the application of these ideas to TD-learning, which will result in the (new) TD-FARL n-tuple algorithm valid for all $N$-player games.

\subsection{TD Learning for $N$ Players} 
\label{sec:TDNTuple3Agt}

\begin{algorithm}[tbp] 
\caption{\textsc{TDFromEpisode}: Perform one episode of TD-learning, starting from state $s_0$. States $s'_{t-1}, s_{t}, s'_t$ and actions $a_{t}$ are for \textbf{one} specific player $p_t$. $r_{t}$ is the delta reward for $p_t$ when taking action $a_t$ in state $s_t$. $A_t$ is the set of actions available in state $s_t$.
}
\label{alg:TdEpisode}
   \begin{algorithmic}[1]
			\Function{TDFromEpisode}{$s_0$} 
			\State $t \leftarrow 0$
			\Repeat 
					\State Choose for player $p_t$ action $a_t \in A_t$ from $s_t$ using policy derived from $V$  
					\State \Comment e.g. $\epsilon$-greedy: with probability $\epsilon$ random, with prob. $1-\epsilon$ using $V$
					\State Take action $a_t$ and observe reward $r_t$, afterstate $s'_t$ and next state $s''$.
					\State $\hat{V}^{new}(s'_{t-1}) = r_t+\gamma V(s'_t)$ \Comment target value for $p_t$'s previous afterstate 
					\State Use NN to get the current value of previous afterstate: $\hat{V}(s'_{t-1})$
					\State Adapt NN by backpropagating error $\delta = \hat{V}^{new}(s'_{t-1}) - \hat{V}(s_{t-1})$
					\State $t \leftarrow t+1$
					\State $s_{t} \leftarrow s''$
			\Until{$s''$ is terminal} 
			\EndFunction
	\end{algorithmic}
\end{algorithm}

\newcommand{\Rnext}{\ensuremath{\vec{r}}\xspace}
\newcommand{\Slast}{\ensuremath{\vec{s}_{last}}\xspace}
\newcommand{\Alast}{\ensuremath{\vec{a}_{last}}\xspace}
\begin{algorithm}[tbp] 
\caption{\textsc{TD-FARL-Episode}: Perform one episode of TD-learning, starting from state $s_0$. Similar to Algorithm~\ref{alg:TdEpisode}, but with Final-Adaptation RL (FARL). 
We connect afterstate $s'$ via player $p_{t}$ with the previous afterstate $\Slast[p_{t}]$ of this player. Note that $\Slast$ and $\Rnext$ are vectors of length $N$. 
}
\label{alg:TD-FARL-Episode}
   \begin{algorithmic}[1]
			\Function{TD-FARL-Episode}{$s_0$} 
			\State $t \leftarrow 0$; 
			\State $\Slast[p] \leftarrow \mbox{null} \quad \forall\, \mbox{player}\, p=0,\ldots,N-1$ \Comment last afterstates
			\Repeat 
					\State $p_{t} =$ player to move in state $s_{t}$
					\State Choose action $a_t$ from $s_t$ using policy derived from $V$  		\Comment e.g. $\epsilon$-greedy
					\State $(\Rnext,s',s'') \leftarrow$ \Call{MakeAction}{$s_t,a_t$} \Comment $s'$: afterstate (after taking $a_t$)
					\State \Comment $\Rnext$ is the delta reward tuple from the perspective of \textbf{all} players $p$
					\State \Call{AdaptAgentV}{$\Slast[p_{t}], \Rnext[p_{t}] , s'$}
					\State $\Slast[p_{t}] \leftarrow s'$  \Comment the afterstate generated by $p_t$ when taking action $a_t$
					\State $t \leftarrow t+1$
					\State $s_{t} \leftarrow s''$  				
			\Until {($s''$ is terminal)}	
			\State \Call{FinalAdaptAgents}{$p_{t}, \Rnext, s'$}	\Comment use final reward tuple to adapt all agents
			\EndFunction
			\State
			\State \Comment Update the  value function (based on NN) for player $p_t$ 
			\Function{AdaptAgentV}{$\Slast[p_{t}], r', s'$}
					\If {($\Slast[p_t] \neq$ null)}			\Comment Adapt $\hat{V}(\Slast[p_t])$ towards target $T$
							\State Target $T 
																= r' + \gamma \hat{V}(s')$ for afterstate $\Slast[p_t]$
							\State Use NN to get $\hat{V}(\Slast[p_t])$  
							\State Adapt NN by backpropagating error $\delta = T - \hat{V}(\Slast[p_t])$
					\EndIf
			\EndFunction
			\State
			\State \Comment Terminal update of value function for all players 
			\Function{FinalAdaptAgents}{$p_{t}, \Rnext, s'$}
				\For {($p=0,\ldots,N-1, \mbox{but } p \neq p_{t}$)}
					\If {($\Slast[p] \neq$ null)}	\Comment Adapt $\hat{V}(\Slast[p])$ towards target $\Rnext[p]$ 
							\State Use NN to get $\hat{V}(\Slast[p])$  
							\State Adapt NN by backpropagating error $\delta = \Rnext[p] - \hat{V}(\Slast[p])$
					\EndIf
				\EndFor
				\State                \Comment Adapt $\hat{V}(s') \rightarrow 0$\,\, ($s'$: terminal afterstate of player $p_t$)
				\State Use NN to get $\hat{V}(s')$   
				\State Adapt NN by backpropagating error $\delta = 0 - \hat{V}(s')$
			\EndFunction
	\end{algorithmic}
\end{algorithm}

We set up a TD-learning algorithm connecting moves to the last move of the \textbf{same} player. This is done in Algorithm~\ref{alg:TdEpisode} (\textsc{TDFromEpisode}). Algorithm~\ref{alg:TdEpisode} shows the TD-learning algorithm in compact form.
It thus makes the general principle clear. But it has the disadvantage that it obscures one important detail: What is shown within the while loop is what has to be done by player $p_t$ in state $s_t$. After completing this, we do however \textit{not} move to the next state $s_{t+1}$ of the same player $p_t$ (one round away), but we let the environment act, get a new state $s''_t$ for the \textbf{next} player, 
and then this next player does \textit{his/her} pass through the while loop. 

To make these details more clear, we write the algorithm down in a form where the pseudocode is closer to the GBG implementation. This is done in Algorithm~\ref{alg:TD-FARL-Episode} (\textsc{TD-FARL-Episode}). Some remarks on Algorithm~\ref{alg:TD-FARL-Episode}:
\begin{itemize}
	\item Now the sequence of states $s_0,s_1,...,s_f$ is really the sequence of consecutive moves in an episode. The players usually vary in cyclic order, $0,1,...,N-1,0,1,...$, but other turn sequences are possible as well. 
	\item In each state the connection to the last afterstate of the same player $p$ is made via $\Slast[p]$. Thus the update step is equivalent to Algorithm~\ref{alg:TdEpisode}.
	\item In contrast to Algorithm~\ref{alg:TdEpisode}, this algorithm has the \FinalAdaptStep\ FARL (function \textsc{FinalAdaptAgents}) included. FARL is covered in more detail in Sec.~\ref{sec:finalAdapt}.
\end{itemize}

Algorithm~\ref{alg:TD-FARL-Episode} is simpler and at the same time more general than our previous TD-algorithms~\citep{Bagh15,Kone15c}
for several reasons:
\begin{enumerate}
	\item Each player has its own value function $V$ and each player seeks actions that \textbf{maximize} this $V$. This is because each $V$ has in its targets the rewards from the perspective of the acting player. So there is no need to set up complicated cases distinguishing between minimization and maximization as it was \citep{Bagh15,Kone15c}. 
	\item The same algorithm is viable for \textbf{arbitrary} 
	number of players.
	\item There is no (or less) unwanted crosstalk because of too frequent updates (as it was the case for some variants in ~\citep{Bagh15,Kone15c}).\footnote{%
	With crosstalk we mean the effect that the update of the value function for one state has detrimental effects on the learned values for other states.}
	\item Since states are connected with states one round (and not one move) earlier, positive or negative rewards propagate back faster. 
\end{enumerate}

These advantages apply as well to SARSA- and Q-learning which we will cover next.

\subsection{SARSA- and Q-Learning for $N$ Players}
\label{sec:SarsaQLearn}


SARSA- and Q-learning have a Q-function instead of the value function in TD-learning. Again,
each player has a Q-function only for \textit{his/her} state-action-pairs $(s_t,a_t)$. The actions of the opponents are subsumed in the reaction from the environment. That is, if $s_t$ is the state for player $p_t$ at time $t$, then $s_{t+1}$ is the next state of the \textit{same} player $p_t$ on which (s)he has to act. For the detailed algorithms (e.g. Algorithm~\ref{alg:Sarsa-FARL-Episode}, 
\textsc{Sarsa-FARL-Episode}) 
the reader is referred to Appendix~\ref{sec:SARSA-Q-Details}. 

\begin{figure}%
\begin{center}
\begin{tikzpicture}	
\matrix (network)
[matrix of nodes,%
nodes in empty cells,
nodes={outer sep=0pt,circle,minimum size=4pt,draw,fill=orange},
column sep={1cm,between origins},
row sep={1cm,between origins}]
{
|[draw=none,fill=none]| & |[xshift=5mm]|& |[draw=none,fill=none]| & |[draw=none,fill=none]| \\
              & 							& 							& 							\\
							& 							& 							& 							\\
};
\foreach \a in {1,...,4}{
	\draw [-stealth] ([yshift=-5mm]network-3-\a.south) -- (network-3-\a); 
	\foreach \b in {1,...,4}{
		\draw (network-3-\a) -- (network-2-\b);}
	\draw (network-1-2) -- (network-2-\a);
}
\draw [-stealth] (network-1-2) -- ([yshift=5mm]network-1-2.north);  
\node[above of =network-1-2,xshift=1mm]{$V(s')$};
\node(st)[below of =network-3-2,xshift=6mm]{afterstate $s'$};
\node[below of =st,yshift=3mm]{(a)};
\end{tikzpicture}
\hspace{3cm}
\begin{tikzpicture}	
\matrix (network)
[matrix of nodes,%
nodes in empty cells,
nodes={outer sep=0pt,circle,minimum size=4pt,draw,fill=orange},
column sep={1cm,between origins},
row sep={1cm,between origins}]
{
|[draw=none,
  fill=none]|  & |[xshift=-4mm]|& |[xshift= 0mm]| & |[xshift=+4mm]| \\
							 & 							  & 							  & 			          &		|[xshift=-6mm]|		\\
|[xshift=+5mm]|& |[xshift=+5mm]|& |[xshift=+5mm]| & |[xshift=+5mm]|\\
};
\foreach \a in {1,...,4}{
	\draw [-stealth] ([yshift=-5mm]network-3-\a.south) -- (network-3-\a); 
	\foreach \b in {1,...,5}{
		\draw (network-3-\a) -- (network-2-\b);}
}
\foreach \b in {1,...,5}{
  \foreach \c in {2,...,4}
		\draw (network-1-\c) -- (network-2-\b);
}
\foreach \c in {2,...,4} {
	\draw [-stealth] (network-1-\c) -- ([yshift=5mm]network-1-\c.north);  
}
\node[above of =network-1-2,xshift=-1mm]{{\small $Q(s,a_1)$}};
\node[above of =network-1-3,xshift= 0mm]{{\small $Q(s,a_2)$}};
\node[above of =network-1-4,xshift= 1mm]{{\small $Q(s,a_3)$}};
\node(st)[below of =network-3-3,xshift=-4mm]{state $s$};
\node[below of =st,yshift=3mm]{(b)};
\end{tikzpicture}	
\end{center}
\vspace{-0.5cm}
\caption{Neural networks in (a) TD-learning, (b) SARSA- and Q-learning.}%
\label{fig:neuralNet}%
\end{figure}
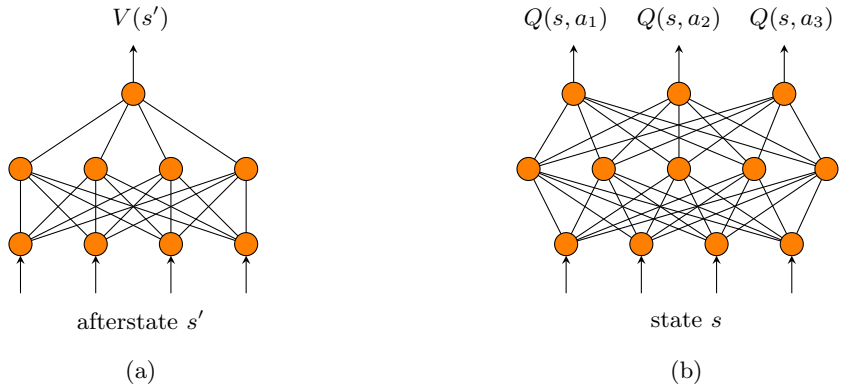

Another key difference is the form of the neural network, as shown in Fig.~\ref{fig:neuralNet}. In TD-learning the network learns $V(s)$ and thus has a single output. It is asked for all afterstates $s'$ arising from available actions as $s'=A(s,a)$ and selects that afterstate that produces the highest $V(s')$. On the other hand, the neural network in Q-learning or SARSA approximates $Q(s,a)$: it has $s$ (or features derived from $s$) as input, but it has \textbf{multiple outputs}, one for each possible action $a$. Given a certain state $s$ it produces a vector of outputs $Q(s,a)$, one element for each $a$. SARSA- or Q-learning selects that action $a$ with the highest output activation. 

SARSA- and Q-learning share with TD-learning the same advantages as mentioned at the end of Sec.~\ref{sec:TDNTuple3Agt}. There are however some disadvantages of SARSA- and Q-learning in games:
\begin{itemize}
	\item If $N_a$ is the number of possible actions, it needs $N_a$ as many weights in the last layer. Since in n-tuple learning there is only one layer, these are $N_a$ as many weights in total.
Consequently, the agents are bigger (in memory and on disk) by a factor $N_a$. This is a considerable increase for larger Hex boards (e.g. $9\times 9$ Hex $\rightarrow N_a=81$) and other games with high branching factor.
	\item SARSA and Q-learning cannot exploit the benefits of afterstates: Several state-action pairs $(s,a)$ may lead to the same afterstate $s'$. Usually the value of an action is fully determined by the resulting afterstate. In such cases, TD-learning stores the value for that afterstate only once, while SARSA- and Q-learning store it multiple times.\footnote{How many times? -- If an afterstate $s'$ contains $x$ pieces of the player who just created $s'$, then there are $x$ preceding states leading with the appropriate action to that same afterstate $s'$. For TicTacToe $x \leq 4$, but for $9\times 9$ Hex $x \leq 40$ (!).}
\end{itemize}
These disadvantages of SARSA- and Q-learning 
are not present in the new TD-learning algorithm~\ref{alg:TD-FARL-Episode}. 

\subsection{Final Adaptation RL (FARL)}
\label{sec:finalAdapt}
Once an episode terminates, we have a delta reward tuple for all players. A drawback of the plain TD- and SARSA-algorithms is that only the current player (who generated the terminal state) uses this information to perform an update step. But the other players can also learn from their (usually negative) rewards. This is what the first part of \textsc{FinalAdaptAgents} (lines 26-28 of Algorithm~\ref{alg:TD-FARL-Episode}) is for: Collect for each player \textit{his} terminal delta reward and use this as target for a final update step where the value of the player's state one round earlier is adapted towards this target.\footnote{%
Why does the target have only the delta reward $\Rnext[p]$ and does not need the value function $\hat{V}(s')$? -- Because the value function for a terminal $s'$ is always 0 (no future rewards are expected).
} 

One might ask whether it is not a contradiction to Sec.~\ref{sec:nPlayer} where we stated that the value for other players is not known for $N>2$. This is not a contradiction: Although intermediate \textit{values} are usual not known for all players, the final \textit{reward} of a game episode -- at least for all games we know of -- \textit{is available} for all players. It is thus a good strategy to use this information for all players.

Second part of \textsc{FinalAdaptAgents}, lines 29-31:
A terminal state is by definition a state where no future rewards are expected. Therefore the value of that state should be zero. However, crosstalk in the network due to the adaptation of other states may lead to non-zero values for terminal states. Jaskowski \cite{jaskowski2018mastering} proposes to make an adaptation step towards target 0 for all terminal states. 
\section{Results}
\label{sec:results}

We show results of our algorithms on two games. In preliminary experiments we tested various settings for parameters, namely the learning rate $\alpha$, the random move rate $\epsilon$ and the eligibility rate $\lambda$. We selected for TicTacToe parameter $\alpha$ linearly decreasing from 1.0 to 0.5 and the n-tuple system consisted just of one 9-tuple. For ConnectFour we used $\alpha=3.7$ and an n-tuple system consisting of initially randomly chosen but then fixed 70 8-tuples. For both games we had $\epsilon$ linearly decreasing from 0.1 to 0.0, $\lambda=0.0$ and we used the TCL scheme as described in~\cite{Bagh15} and Appendix~\ref{sec:TCL}. Note that due to TCL the effective learning rate adopted by most weights can be far smaller than $\alpha$.

\subsection{TicTacToe} 
\begin{figure}%
\includegraphics[width=0.5\columnwidth]{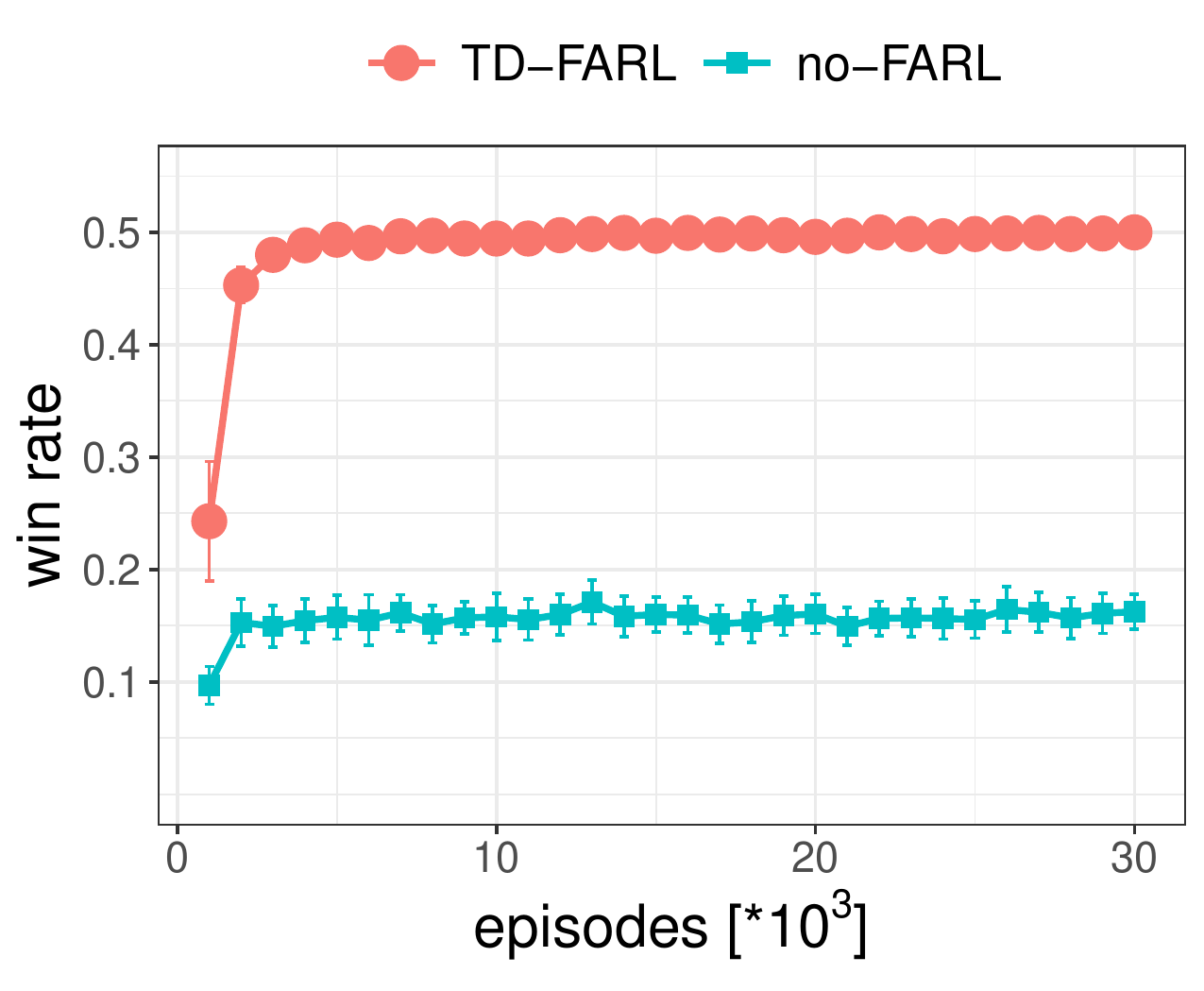}%
\includegraphics[width=0.5\columnwidth]{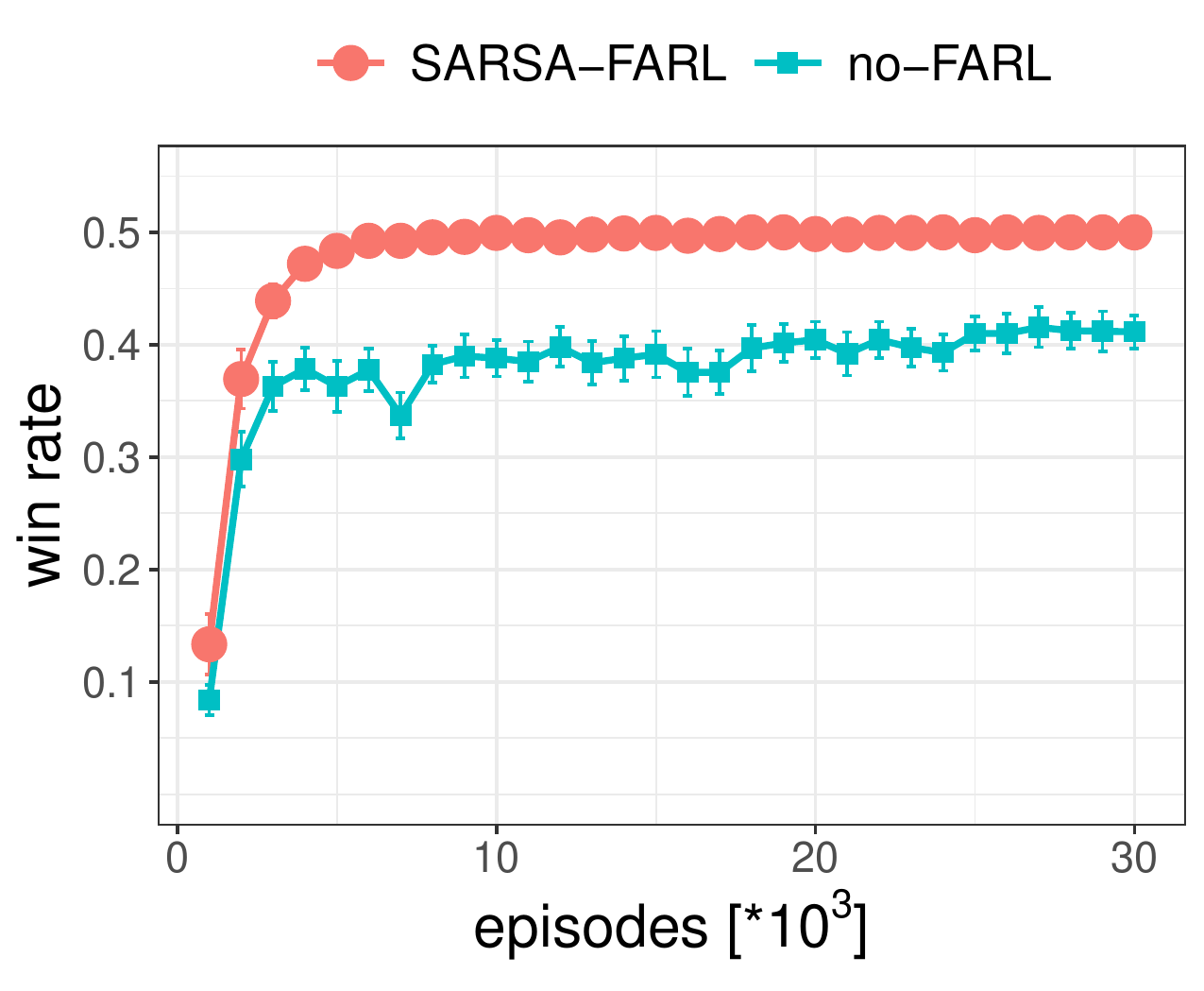}%

\caption{Different versions of TD-learning (left) and SARSA-learning (right) on TicTacToe. Each agent is evaluated by playing games from different start positions in both roles, 1\textsuperscript{st} and 2\textsuperscript{nd} player, against the perfect-playing Max-N agent. The best achievable result is 0.0, because Max-N will win at least in one of the both roles. Shown is the average over 25 training runs.
}%
\label{fig:TD-SARSA-TTT}%
\end{figure}

Fig.~\ref{fig:TD-SARSA-TTT} shows learning curves of TD- and SARSA-learning. 
The red curves show the full Algorithms~\ref{alg:TD-FARL-Episode} (\textsc{TD-FARL-Episode}) and \ref{alg:Sarsa-FARL-Episode} 
(\textsc{Sarsa-FARL-Episode}). Algorithm~\ref{alg:TD-FARL-Episode} is slightly quicker in convergence, but both algorithms will eventually master the simple game TicTacToe. 
The blue curves show the results when we switch off \textsc{FinalAdaptAgents}: The decrease in performance is severe: drastic for TD and strong for SARSA. 


\subsection{ConnectFour} 
\label{sec:result-C4}

\begin{figure}[ht]%
\centerline{
%
\includegraphics[width=0.9\columnwidth]{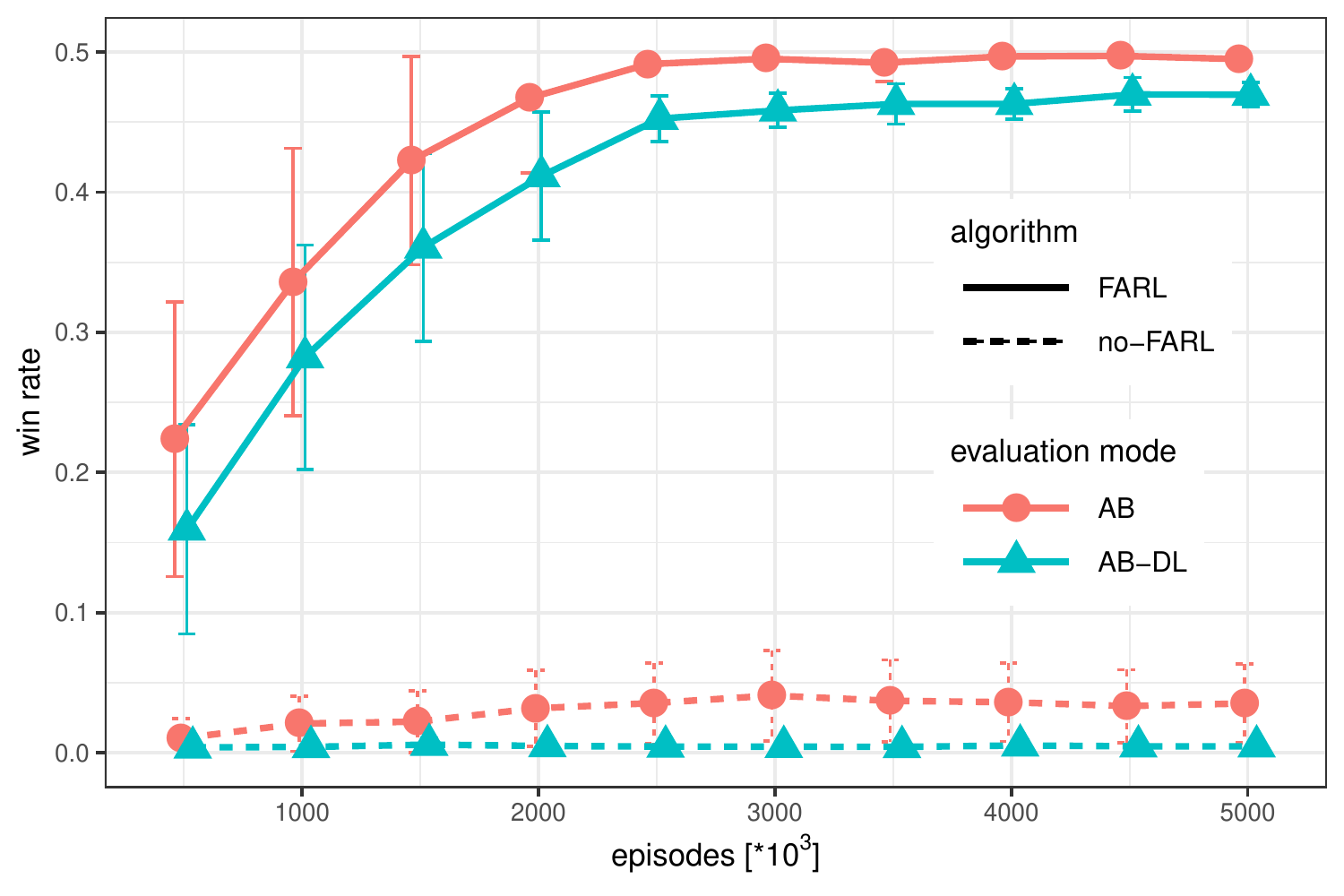}}

\caption{
TD-learning on ConnectFour. During training, agent TD-FARL is evaluated against the perfect-playing agents AlphaBeta (AB) and AlphaBeta-with-distant-losses (AB-DL). Both agents play in both roles (first or second). Since ConnectFour is a theoretical win for the starting player, the ideal win rates against AB and AB-DL are 0.5. The solid lines show the mean win rates from 10 training runs with FARL. 
The dashed curves \textit{no-FARL} show the results when FARL is turned off. Error bars depict the standard deviation of the mean.
}
\label{fig:TDNT3-C4}%
\end{figure}

Fig.~\ref{fig:TDNT3-C4} shows learning curves of our TD-FARL agent on the non-trivial game ConnectFour. 
Two modes of evaluation are shown: The red curves evaluate against opponent AlphaBeta (AB), the blue curves against opponent AlphaBeta-Distant-Losses (AB-DL). 
The AlphaBeta algorithm extends the Minimax algorithm by efficiently pruning the search tree. Thill et al.~\cite{Thil14} were able to implement AlphaBeta for  ConnectFour in such a way that it plays perfect in situations where it can win. AB and AB-DL  differ in the way they react on losing states: While AB just takes a random move, AB-DL searches for the move which postpones the loss as far (distant) as possible. It is tougher to win against AB-DL since it 
will punish every wrong move. The final results for our TD-FARL agent are however very satisfying: 49.5\% win rate against AB, 46.5\% win rate against AB-DL. It is worth noting that two perfect-playing opponents (AB and AB-DL) are not necessarily equally strong.


It is a remarkable success that TD-FARL
learns only from training by self-play  to defeat the perfect-playing AlphaBeta agents in 49\%/46\% of the cases. Remember that TD-FARL has never seen AlphaBeta before during training. The result is similar to our previous work~\cite{Bagh15}. But the difference is that the new algorithm can be applied without any change to other games with any $N$. 

There is also a striking failure visible in Fig.~\ref{fig:TDNT3-C4}: If we switch off \textsc{FinalAdapt\-Agents} (curves \textit{no-FARL}), we see a complete break-down of the TD agent: It loses nearly all its games. We conclude that the part propagating the final reward of the other player back to the other player's previous state is vitally important.\footnote{%
It is really the first part of FARL which is important: We conducted an experiment where we switch off only the second part of FARL and observed only a very slight degradation (1\% or less).}
If we analyze the \textit{no-FARL}-agent we find that it has only 0.9\% active weights while the good-working TD-FARL agent has 8\% active weights. This comes because the other player (that is the one who loses the game since the current player created a winning state) has never the negative reward propagated back to previous states of that other player. Thus the network fails to learn threatening positions and/or precursors of such threatening positions.


\begin{table}[tb]
    \caption{Results for Algorithm~\ref{alg:TD-FARL-Episode} (\textsc{TD-FARL-Episode}) on various games. In Nim(3P) $h$x$s$, there are initially $h$ heaps with $s$ stones. For each game, 10 training runs with different seeds are performed and the resulting TD agent is evaluated by playing against opponents as indicated in column 3 (two such opponents in the case of Nim3P). Each agent plays all roles. 
    Shown are the TD agent's win rates or scores (rewards): mean from 10 runs plus/minus one standard deviation of the mean.
    } 
    \label{tab:allGames}
    \centering
    \begin{footnotesize}
    \begin{tabular}{|l|l|l|c|c||c|c|} \hline\hline
     \multirow{2}{*}{Game} & \multirow{2}{*}{N}
                 & evaluated & \multicolumn{2}{|c||}{win rates or scores}     
                 & \multicolumn{2}{|c|}{\multirow{2}{*}{other RL research}}  \\ \cline{4-5}
             &   & vs.        & FARL             &  no-FARL  & \multicolumn{2}{|c|}{}   \\ \hline \hline
    2048 & 1 &           & $142\,000\pm 1\,000$& $122\,000\pm 900$ 
    & \cite{jaskowski2018mastering} & $80\,000$ \\ \hline
    TicTacToe& 2 & Max-N$_{10}$ \cite{korf1991maxN}& $49\%\pm 5\%$    & $18\%\pm 6\%$ & \multicolumn{2}{|c}{}   \\ \hline
    \multirow{2}{*}{ConnectFour} & \multirow{2}{*}{2}
                   & AB \cite{Thil14}      & $49.5\%\pm 0.5\%$& $3.5\%\pm 0.1\%$ 
                   & \cite{dawson2020} & $0.0\% \pm 0.0\%$ \\ \cline{3-7}
               &   & AB-DL \cite{Thil14}   & $46.5\%\pm 0.5\%$& $0.0\%\pm 0.1\%$ 
               & \multicolumn{2}{|c}{} \\ \cline{1-5}
    Hex 6x6& 2& MCTS$_{10000}$& $81\%\pm 5\%$    & $0.0\%\pm 0.2\%$ 
    & \multicolumn{2}{|c}{} \\ \cline{1-5}
    \multirow{2}{*}{Othello} & \multirow{2}{*}{2}
                & Edax$_{d1}$ \cite{delorme2019}& $55\%\pm 1\%$    & $53\%\pm 1\%$   & \multicolumn{2}{|c}{} \\ \cline{3-7}
           &    & \textsc{Bench} \cite{Ree2013reinforcement}      & $95\%\pm 0.3\%$  & $96\%\pm 0.2\%$  & \cite{Ree2013reinforcement} & $87.1\%\pm 0.9\%$\\ \hline
    Nim 3x5 & 2 & Max-N$_{15}$ \cite{korf1991maxN}& $50\%\pm 1\%$    & $12\%\pm 6\%$ & \multicolumn{2}{|c}{}   \\ \cline{1-5}
    \multirow{2}{*}{Nim3P 3x5} & \multirow{2}{*}{3}
                & Max-N$_{15}$ \cite{korf1991maxN}& $0.33\pm 0.03$    & $0.03\pm 0.01$   & \multicolumn{2}{|c}{} \\ \cline{3-5}
            &   & MCTS$_{5000}$ & $0.78\pm 0.02$    & $0.09\pm 0.02$  
            &  \multicolumn{2}{|c}{}  \\ \hline\hline
    \end{tabular}
    \end{footnotesize}
\end{table}

\subsection{A Variety of Games}
In Table~\ref{tab:allGames} we show the results for seven  games with varying number of players (1, 2, or 3). While there exist many well-known games for 1 and 2 players, it is not easy to find 3-player games which have a clear winning strategy. 
Nim3P, the 3-player-variant of the game Nim, is such a game. Each player can take any number of pieces from one heap at his/her turn. The player who takes the last piece loses and gets a reward of 0.0, then 
the successor is the winner and gets a reward of 1; the predecessor gets a reward of 0.2. This smaller reward helps to break ties in otherwise 'undecided' situations.  
The goal for each player is to maximize his/her average reward.
Nim3P cannot end in a tie.

All games are learned by exactly the same TD-FARL / no-FARL algorithm. The strength of the resulting agent is evaluated by playing against opponents, where all agents play in all roles. The opponents are in many cases  perfect-playing or strong-playing agents. If all agents play perfect, the best possible result for each agent is a win rate of 50\% for 2-player games and a score of 0.4 for the game Nim3P (one third of the total reward 1.2 distributed in each episode). Max-N$_d$ is an N-Player tree search with depth $d$~\cite{korf1991maxN}, being a perfect player for the games TicTacToe, Nim, Nim3P. For ConnectFour, AB and AB-DL~\cite{Thil14} are perfect-playing agents introduced in Section~\ref{sec:result-C4}.  Edax$_{d1}$~\cite{delorme2019} is a strong Othello program, played here with depth 1. \textsc{Bench}~\cite{Ree2013reinforcement} is a medium-strength Othello agent. MCTS$_a$ is a Monte Carlo Tree Search with $a$ iterations.

As can be seen from Table~\ref{tab:allGames}, TD-FARL reaches near-perfect playing strength in most competitions against (near-)perfect opponents and it dominates non-perfect opponents. The most striking feature of Table~\ref{tab:allGames} is its column 'no-FARL': it is in all games much weaker, with one notable exception: In Othello the results for TD-FARL and TD-no-FARL are approximately the same. This is supported by the results from van der Ree and Wiering~\cite{Ree2013reinforcement} who had good results on Othello with their no-FARL algorithms. We have no clear answer yet why Othello behaves differently than all other games. 

\subsection{Comparison with Other RL Research}
\label{sec:comparisonRL}
For some games we compare in Table~\ref{tab:allGames} with other RL approaches from the literature. \Jaskowski~\cite{jaskowski2018mastering} achieves for the game 2048 with a similar amount of training episodes and a general-purpose baseline TD agent scores around $80\,000$. It has to be noted that \Jaskowski~with ten times more training episodes and algorithms specifically designed for 2048 reaches much higher scores around $600\,000$, but here we only want to compare with general-purpose RL approaches. 

Dawson~\cite{dawson2020} introduces a CNN-based and AlphaZero-inspired~\cite{silver2017AlphaGoZero} RL agent named ConnectZero for the game ConnectFour, which can be played online. Although it reaches a good playing strength against MCTS$_{1000}$, it cannot win a single game against our AlphaBeta agent. We performed 10 episodes with ConnectZero starting (which is a theoretical win), but found that instead AlphaBeta playing second won all games. This is in contrast to our TD-FARL which wins nearly all episodes when starting against AlphaBeta. 

Finally we compare for the game Othello with the work of van der Ree and Wiering~\cite{Ree2013reinforcement}: Their Q-learning agent reaches against \textsc{Bench} (positional player) a win rate of 87\% while their TD-learning agent reaches 72\%. Both win rates are a bit lower than our 95\%.

\subsection{Discussion}
Looking at the results for ConnectFour, one might ask the following question: If FARL is so important for RL-based ConnectFour, why could Bagheri et al.~\cite{Bagh15} learn the game when their algorithm did not have FARL? -- The reason is, that both algorithms have different TD-learning schemes: While the algorithm in \cite{Bagh15} propagates the target from the current state back to the previous state (one \textit{move} earlier), our $N$-player RL propagates the target from the current state back to the previous move of the same player (one \textit{round} earlier). The $N$-player FARL  is more general (it works for arbitrary $N$). But it has also this consequence: If for example a 2-player game is terminated by a move of player 1, the value of the previous state $\Slast[p_2]$ of player 2 is never updated. As a consequence, player 2 will never learn to avoid the state preceding its loss. Exactly this is cured, if we activate FARL.


\section{Conclusion and Future Work}
In summary, we collected evidence that Algorithm~\ref{alg:TD-FARL-Episode} (\textsc{TD-FARL-Episode}) 
produces good results on a variety of games.  It has been shown that the new ingredient FARL (the final adaption step) is vital in many games to get these good results.
%

Compared to~\cite{Bagh15}, TD-FARL has the benefit that it can be applied unchanged to all kind of games whether they have one, two or three players. The algorithm of \cite{Bagh15} cannot be applied to games with more than two players. 


We see the following lines of direction for future work: (a) More 3-player games. Although Nim3P with a clear winning strategy provided a viable testbed for evaluating our algorithm, taking more 3-player or $N$-player games into account will help us to investigate how well our introduced methods generalize. 
(b) Can we better understand why Othello is indifferent to using FARL or no-FARL? Are there more such games? If so, an interesting research question would be whether it is possible to identify common game characteristics that allow to decide whether FARL is important for a game or not.

%


\newpage
\appendix
\section{Appendix}

\begin{table}%
\caption{Symbols and variables}
\label{tab:symbVar}
\noindent
\begin{tabular}{l|p{10.5cm}}
  \hline
	symbol		& text \\
	\hline
	$s_t$ 		& game state at time $t$ \\
	$a_t$			& action taken at time $t$ \\
	$A_t$			& set of actions available in state $s_t$ \\
	$p_t$ 		& $p_t \in \{0,\ldots,N-1\}$: player to move in state $s_t$  \\
	$N$				& number of players in the game \\
	$r_{t+1}$ & reward delta when arriving with $(s_t,a_t)$ in state $s_{t+1}$ \\
	$R_{t+1}$ & absolute (cumulative) reward in state $s_{t+1}$ \\
	$T_{lrnRM}$  & boolean switch: If set to true, then TD-update shall be done after a random action (see line 2 of Algorithm~\ref{alg:tdlambda}) \\
	$\epsilon \in [0,1]$ & random move rate \\
	$\gamma \in [0,1] $	 & discount factor, usually 1.0 
\end{tabular}
\end{table}

\subsection{Nomenclature}
\label{sec:prelim}

We list in Table~\ref{tab:symbVar} the symbols and variables used in this paper. 

\subsection{SARSA and Q-Learning: Detailed Algorithms}
\label{sec:SARSA-Q-Details}

\begin{algorithm}[tbp] 
\caption{\textsc{QLearnFromEpisode}: Perform one episode of Q-learning, starting from state $s_0$. States $s_{t}$ and actions $a_{t}$ are for \textbf{one} specific player $p$. $r_{t+1}$ is the delta reward for $p$ when taking action $a_t$ in state $s_t$. $A_t$ is the set of actions available in state $s_t$. Additional inputs are:  
$\gamma$ (discount factor), and $\epsilon$ (random move rate).}
\label{alg:qlearnEpisode}
   \begin{algorithmic}[1]
			\Function{QLearnFromEpisode}{$s_0$} 
			\State $t \leftarrow 0$
			\While {true} 
					\State Choose action $a_t \in A_t$ from $s_t$ using policy derived from $Q$  
					\State \Comment e.g. $\epsilon$-greedy: with probability $\epsilon$ random, with prob. $1-\epsilon$ using $Q$
					\State Execute $a_t$ and observe $r_{t+1}, s_{t+1}$ 		\Comment after environment and opponents act 
					\State Use NN to get $\hat{Q}(s_{t+1},a') \quad\forall a' \in A_{t+1}$
					\State $ M \leftarrow \stackrel[a' \in A_{t+1}]{}{\max} \hat{Q}(s_{t+1},a')$ \Comment if $s_{t+1}$ is terminal, set $M=0$
					\State Target $\hat{Q}^{new}(s_{t},a_t) = r_{t+1} + \gamma M$
					\State Use NN to get $\hat{Q}(s_{t},a_t)$  \Comment only \textbf{one} action $a'$
					\State Adapt NN by backpropagating error $\delta = \hat{Q}^{new}(s_{t},a_t) - \hat{Q}(s_{t},a_t)$
					\State \Comment NN-outputs for all actions other than $a_t$ have their $\delta$ set to 0
					\If{$s_{t+1}$ is terminal} \textbf{break} \Comment out of while \EndIf
					\State $t \leftarrow t+1$
			\EndWhile		
			\EndFunction
	\end{algorithmic}
\end{algorithm}

\begin{algorithm}[tbp] 
\caption{\textsc{SarsaFromEpisode}: Perform one episode of SARSA-learning, starting from state $s_0$. States $s_{t}$ and actions $a_{t}$ are for \textbf{one} specific player $p$. $r_{t+1}$ is the delta reward for $p$ when taking action $a_t$ in state $s_t$. $A_t$ is the set of actions available in state $s_t$. Additional inputs are: $\gamma$ (discount factor), and $\epsilon$ (random move rate).}
\label{alg:sarsaEpisode}
   \begin{algorithmic}[1]
			\Function{SarsaFromEpisode}{$s_0$} 
			\State $t \leftarrow 0$
			\While {true} 
					\State Choose action $a_t \in A_t$ from $s_t$ using policy derived from $Q$  
					\State \Comment e.g. $\epsilon$-greedy: with probability $\epsilon$ random, with prob. $1-\epsilon$ using $Q$
					\State Execute $a_t$ and observe $r_{t+1}, s_{t+1}$ 		\Comment after environment and opponents act 
					\State Choose action $a_{t+1} \in A_{t+1}$ from $s_{t+1}$ using policy derived from $Q$  
					\State $ M \leftarrow  Q(s_{t+1},a_{t+1})$ 			\Comment if $s_{t+1}$ is terminal, set $M=0$
					\State Target $\hat{Q}^{new}(s_{t},a_t) = r_{t+1} + \gamma M$
					\State Use NN to get $\hat{Q}(s_{t},a_t)$  \Comment only \textbf{one} action $a_t$
					\State Adapt NN by backpropagating error $\delta = \hat{Q}^{new}(s_{t},a_t) - \hat{Q}(s_{t},a_t)$
					\State \Comment NN-outputs for all actions other than $a_t$ have their $\delta$ set to 0
					\If{$s_{t+1}$ is terminal} \textbf{break} \Comment out of while \EndIf
					\State $t \leftarrow t+1$
			\EndWhile		
			\EndFunction
	\end{algorithmic}
\end{algorithm}


\newcommand{\pnext}{\ensuremath{p_{next}}\xspace}
\begin{algorithm}[tbp] 
\caption{\textsc{Sarsa-FARL-Episode}: Equivalent to \textsc{SarsaFromEpisode}, but the pseudocode is closer to the GBG implementation. $s''$ is the next state from $s_t$ after taking action $a_t$, usually with another player $\pnext$ to move. We connect it via $\pnext$ with the previous state $\Slast[\pnext]$ of this player. Note that $\Slast, \Alast$ are vectors of length $N$. }
\label{alg:Sarsa-FARL-Episode}
   \begin{algorithmic}[1]
			\Function{Sarsa-FARL-Episode}{$s_0$} 
			\State $t \leftarrow 0$    
			\State Choose action $a_0$ from $s_0$ using policy derived from $Q$  \Comment e.g. $\epsilon$-greedy, using NN
			\State $\Slast[p] \leftarrow (p=p_{0}) \,?\, s_{0} : \mbox{null} \quad \forall\, \mbox{player}\, p=0,\ldots,N-1$
			\State $\Alast[p] \leftarrow (p=p_{0}) \,?\, a_{0} : \mbox{null} \quad \forall\, \mbox{player}\, p=0,\ldots,N-1$
			\Repeat 
					\State $(\Rnext,s',s'') \leftarrow$ \Call{MakeAction}{$s_t,a_t$} \Comment afterstate $s'$ is not used
					\State \Comment $\Rnext$ is the delta reward tuple from the perspective of \textbf{all} players $p$
					\State $\pnext =$ player to move in next state $s''$
					\State \Call{AdaptAgentQ}{$\pnext, \Rnext[\pnext], s''$}
					\State Choose action $a''$ from $s''$ using policy derived from $Q$  
					\State \Comment e.g. $\epsilon$-greedy, using NN. Calculate $a''$ anew after \Call{AdaptAgentQ} (!)
					\State $t \leftarrow t+1$
					\State $\Slast[\pnext] \leftarrow s_{t} \leftarrow s''$
					\State $\Alast[\pnext] \leftarrow a_{t} \leftarrow a''$
			\Until {($s''$ is terminal)}	
			\State \Call{FinalAdaptAgents}{$\pnext, \Rnext$}	\Comment use final reward tuple to adapt other agents
			\EndFunction 
			\State 
			\State \Comment Update the Q-function (based on NN) for player $p$ to act in $s''$
			\Function{AdaptAgentQ}{$p, r'', s''$}
					\If {($s''$ is terminal)}
							\State $M=0$
					\Else
							\State Choose action $a''$ from $s''$ using $Q$-derived policy   \Comment e.g. $\epsilon$-greedy, using NN
							\State $ M \leftarrow  Q(s'',a'')$ 			
					\EndIf
					\If {($\Slast[p] \neq$ null)}			\Comment Adapt $\hat{Q}$ towards target $T$
							\State Target $T = 
																 r'' + \gamma M$
							\State Use NN to get $\hat{Q}(\Slast[p],\Alast[p])$  \Comment only \textbf{one} action 
							\State Adapt NN by backpropagating error $\delta = T - \hat{Q}(\Slast[p],\Alast[p])$
							\State \Comment NN-outputs for all actions other than $\Alast[p]$ have their $\delta$ set to 0
					\EndIf
			\EndFunction
			\State
			\State \Comment Terminal update of Q-functions for all players \textbf{other} than $\pnext$
			\Function{FinalAdaptAgents}{$\pnext, \Rnext$}
				\For {($p=0,\ldots,N-1, \mbox{but } p \neq \pnext$)}
					\If {($\Slast[p] \neq$ null)}  \Comment Adapt $\hat{Q}$ towards target $T = \Rnext[p]$		
							\State Use NN to get $\hat{Q}(\Slast[p],\Alast[p])$  \Comment only \textbf{one} action 
							\State Adapt NN by backpropagating error $\delta = \Rnext[p] - \hat{Q}(\Slast[p],\Alast[p])$
					\EndIf
				\EndFor
			\EndFunction
	\end{algorithmic}
\end{algorithm}

Algorithm~\ref{alg:qlearnEpisode} and \ref{alg:sarsaEpisode} show Q-learning and SARSA algorithms in their shortest form, close to \citep{Ree2013reinforcement,SuttBart98}, solely from the perspective of one player. $s_t$ and $s_{t+1}$ are one \textit{round} away (same player, next round). These algorithms are quite compact, but again they omit one detail: After a pass through the while loop with state $s_t$, the next pass is \textit{not} with state $s_{t+1}$ (same player, next round), but with the state $s''$ resulting from advancing $s_t$ with $a_t$. Usually, $s''$ has another player $p'' = (p_t+1)\% N$.

It is tempting, but dangerous (or plain wrong) to try to work in Algorithm~\ref{alg:sarsaEpisode} with only \textit{one} 'Choose action $a_t$' per training cycle: Could we not re-use the $a_{t+1}$ obtained in every pass as the $a_t$ of the next pass through the while loop? - The answer is \textit{No} because after choosing $a_{t+1}$ we change $Q$, and thus the action obtained from $s_{t+1}$ may be different.

Algorithm~\ref{alg:Sarsa-FARL-Episode} shows the SARSA algorithm in a somewhat longer form exposing this and other details. It is closer to the GBG-implementation and it is for the case that we have one NN (the n-tuple network) used by all players and that training proceeds by having all SARSA agents play against themselves. Now $s_t$ and $s_{t+1}$ are one \textit{move} away (next player). We store in $\Slast[p]$ the last state of the same player $p$ one round earlier and thus make the update step equivalent to Algorithm~\ref{alg:sarsaEpisode}. 

There is with \textsc{FinalAdaptAgents} one more important difference to algorithm~\ref{alg:sarsaEpisode}: Once an episode is finished ($s_{t+1}$ is terminal), we add one fictious round where no moves are made but all players $p$ perform an update towards their final reward $\Rnext[p]$. This speeds up learning, because the positive reward from a player making a winning move is directly propagated to its previous state.\footnote{%
The second part of \textsc{FinalAdaptAgents} in Algorithm~\ref{alg:TD-FARL-Episode}, where the final $\hat{V}$ is adapted towards 0, is not needed in Algorithm~\ref{alg:Sarsa-FARL-Episode}, because the $\hat{Q}$-value of a terminal state is never used in SARSA.
}


In Sec.~\ref{sec:SarsaQLearn} we stated that SARSA and Q-learning cannot benefit from the usage of afterstates. 
	Why is it not possible to use in $Q(s,a)$ afterstates $s'$ instead of states $s$? -- When these algorithm have to decide which action to take, they cannot rely on the afterstate $s'$, but they have to use the next state $s''$. This is because the random element added by the environment may have important influence on the action to take.

\hypertarget{hrefEligibility}{\subsection{Eligibility Method}}
\label{sec:eligibility}
The general \textsc{TD($\lambda$)Update} rule for weights $w_t$ is \citep{SuttBart98}: 
\begin{eqnarray}
		w_{t+1} &=& w_t + \alpha\delta_t \sum_{k=t_0}^{t}{\lambda^{t-k}\nabla_{w_k} V_{w_k}(s_k)} \nonumber \\
					  &=& w_t + \alpha\delta_t \nabla_{w_t} V_{w_t}(s_t) \label{eq:eligibility} \\
					  & & \hphantom{w_t} + \alpha\delta_t \nabla_{w_{t-1}} V_{w_{t-1}}(s_{t-1}) \lambda \nonumber\\
					  & & \hphantom{w_t} + \alpha\delta_t \nabla_{w_{t-2}} V_{w_{t-2}}(s_{t-2}) \lambda^2 \nonumber\\
						& & \hphantom{w_t} + \cdots \nonumber
\end{eqnarray}
where in the last equation the sum is written out from index $t$ \textit{downwards} to index $t_0$.

Normally, this formula is realized in TD($\lambda$) algorithms with the help of eligibility traces~(\cite{Kone15c,SuttBart98,Thil14}) and $t_0=0$ is used. But eligibility traces are unpractical (prohibitively slow) for games like 2048 (long episodes, millions of weights would need repeated updates\footnote{%
As Jaskowski~\cite{jaskowski2018mastering} writes: \glqq The eligibility traces vector is of the same length as the weight vector. The standard Sutton’s implementation consists in updating all the elements of the vector at each learning step. This is computationally infeasible when dealing with hundreds of millions of weights.\grqq\ Thill~\cite{Thil14} reduced this to the number of weights \textit{activated} during an episode. Although this is sensible for ConnectFour (at most 42 moves; it was shown that each episode would not activate more than approximately 5000 weights), it is not viable for 2048: Since in 2048 an episode can last for several 10.000 moves, this may activate nearly all eligibility traces in the network. This would make an update step prohibitively slow.}).

Jaskowski~\cite{jaskowski2018mastering} has the brilliant idea to use a finite horizon $t_0=\max(t-h, 0)$,  
that is, we use from the TD($\lambda$)-equation~\eqref{eq:eligibility} only the first $h+1$ terms.\footnote{%
In the beginning, if $t<h$, it might be even less terms.}
 If we choose $h=\left\lfloor log_{\lambda}(0.1)\right\rfloor$ then we retain only those terms with $\lambda^{t-k} \geq 0.1$. This realizes a constant and relatively small computational effort, irrespective of how long an episode is. Note that episodes in the game 2048 can often have more than 10.000 moves.\footnote{%
To build in the game 2048 a \glqq 16384\grqq\ tile, we need at least 16384/2 = 8092 tiles \glqq 2\grqq. In 90\% of the moves we get a new tile \glqq 2\grqq\ (only 10\% of the moves have a new tile \glqq 4\grqq). Thus, 8192 moves is a rough estimate for an episode ending with a \glqq 16384\grqq-tile.}

For n-tuple systems the value function in afterstate $s'$ (for player $p(s')$) is 
\begin{equation}
			V(s') = \sigma\left( \sum_{i=1}^m{ \sum_{q \in S(s')}{w_i[Ind_i(q)]}}\right) =: \sigma\left( \nu(s') \right)
\label{eq:valueNtuple}
\end{equation}
\noindent
with $S(s')$: the set of all states symmetric to $s'$ (including $s'$ itself), $N_S = |S(s')|$,  
$Ind_i(q)$: the index into the look-up table $w_i$ of n-tuple $i$ given state $q$, and $m$ being the number of n-tuples. The term $\nu(s')$ (the double sum) is the activation of the n-tuple network induced by state $s'$.

If we take in Eq.~\eqref{eq:valueNtuple} the derivative of $V(s')$ w.r.t. weight $w_j$, it is either 0  (if $w_j$ is not activated by the current state $s'$) or it is 
$$
			\frac{\partial}{\partial w_j} V(s') = \sigma'\left( \nu(s') \right)\cdot 1
$$
if $w_j$ is activated by the current state $s'$. 
Thus the update function Eq.~\eqref{eq:eligibility} reduces in the special case $\lambda=0$ and $V(s')$ according to Eq.~\eqref{eq:valueNtuple}  to 
\begin{equation}
		w_i[Ind_i(q)] = w_i[Ind_i(q)] + \frac{1}{mN_S}\alpha\delta_t\sigma'\left(\nu(s')\right) \quad\mbox{with } q \in S(s')
\label{eq:v_update}
\end{equation}
\noindent
with error signal $\delta_t = T - V(s')$, target $T$ and $\sigma'\left(\nu(s')\right)$ being the derivative of sigmoid $\sigma()$ w.r.t. the activation induced by afterstate $s'$.\footnote{This means $\sigma'\left(\nu(s')\right)=1-V^2(s')$ in the case of $\sigma=\tanh$ and $\sigma'\left(\nu(s')\right)=1$ in the case of the identity function $\sigma=\mbox{\textit{id}}$.} 
%
The new scaling factor $1/(m N_S)$, introduced by~\cite{jaskowski2018mastering}, ensures that for a linear net ($\sigma=\mbox{\textit{id}}$) an update with $\alpha=1$ moves $V(s')$ directly to the target value $T$, irrespective of the number of n-tuples and number of equivalent states.\footnote{%
This can be shown by inserting Eq.~\eqref{eq:v_update} into Eq.~\eqref{eq:valueNtuple}:
\begin{eqnarray}
		V^{(new)}(s') 
							&=& \sum_{i=1}^m{ \sum_{q \in S(s')}{\left( w_i[Ind_i(q)] + \frac{\alpha}{mN_S}\left(T-V(s')\right) \right)}} \\
							&=& \sum_{i=1}^m{ \sum_{q \in S(s')}{\left( w_i[Ind_i(q)] \vphantom{\frac{1}{mN_S}}\right)}} 
									+ mN_S \frac{\alpha}{mN_S}\left(T-V(s')\right) \nonumber\\
							&=& V(s') + \alpha \left(T-V(s')\right) \nonumber\\
							&=& \alpha T + (1-\alpha) V(s') \quad\stackrel[\alpha \rightarrow 1]{}{\longrightarrow}\quad T  \nonumber
\label{eq:Vscaling}
\end{eqnarray}
} 
	
During an update of equivalent states, each n-tuple index is updated not more than once. See Sec.~\ref{sec:helper} for more details.

The algorithm \textsc{TD($\lambda$)Update} implementing this is shown below in Algorithm~\ref{alg:tdlambda}. 

\subsection{Helper Algorithms~\ref{alg:update} (\textsc{MakeAction}) and \ref{alg:tdlambda} (\textsc{TD($\lambda$)Update})}
\label{sec:helper}

Some remarks on the helper algorithms \textsc{MakeAction} and \textsc{TD($\lambda$)Update}:
\begin{itemize}
	\item \textsc{MakeAction} is called once per loop in Algorithms~\ref{alg:TD-FARL-Episode} and \ref{alg:Sarsa-FARL-Episode} above.
	\item \textsc{TD($\lambda$)Update}$(s'_t,\delta,0)$ is called each time we write \textit{'Adapt NN by backpropagating error $\delta \ldots$'} in the Algorithms~\ref{alg:TD-FARL-Episode} and \ref{alg:Sarsa-FARL-Episode} above.
	\item The reason for the index set $M$ in lines 5, 7 and 9 of Algorithm~\ref{alg:tdlambda} is as follows: Each state $s' \in S(s'_k)$ activates exactly one weight $w_i[Ind_i(s')]"$ in each n-tuple. But multiple equivalent states $s'$ may activate the same weight $w_i$ repeatedly. This may lead to a too large update step in line 8 of Algorithm~\ref{alg:tdlambda}. To avoid this, we keep in $M$ a set of already visited indices $Ind_i(s')$ and skip the update step, if the weight was updated before. 
\end{itemize}

The relevant classes implementing the new TD and SARSA algorithms are \texttt{TDNTuple3Agt} and \texttt{SarsaAgt}, resp.\footnote{%
The older class \texttt{TDNTuple2Agt}~\cite{Bagh15,Kone15c}  is now deprecated.}

Some elements of the algorithms and their counter part in the GBG source code
 are listed in Table~\ref{tab:symbSource}.\\[0.1cm]

\begin{algorithm}[tbp] 
\caption{\textsc{MakeAction}: Take action $a$ in state $s$ and observe cumulative reward tuple $R$, \afterstate\ $s'$ and next state $s''$.}
\label{alg:update}
   \begin{algorithmic}[1]
			\Function{MakeAction}{$s, a$} 
			\State $s' \leftarrow$ \Call{ComputeAfterState}{$s,a$}	
			\State $(s'',R) \leftarrow$ \Call{AddRandomPart}{$s'$}				\Comment $R=\left(R(s''|p)\mid p=0,\ldots,N-1\right)$
			\If {!\,$T_{after}$}
					\State $s' \leftarrow s''$	\Comment	afterstate and next state shall be the same 	
			\EndIf
			\State \Return $(R,s',s'')$
			\EndFunction
	\end{algorithmic}
\end{algorithm}

\begin{algorithm}[tbp] 
\caption{\textsc{TD($\lambda$)Update}. Inputs:  $\delta_t$ is the delta signal for the actual afterstate $s'_t$. 
Additional inputs are: $T_{lrnRM}$ (see Table~\ref{tab:symbVar}), step size $\alpha$, eligibility rate $\lambda$, number $m$ of n-tuples, horizon $h$, and the  $h$ afterstates preceding $s'_t$. The variables $S(s'_k)$, $N_S$, $\sigma'\left(\nu(s'_k)\right)$ and $Ind_i(s')$ are explained in Sec.~\ref{sec:eligibility} (text around Eqs.~\eqref{eq:valueNtuple} and \eqref{eq:v_update}).}
\label{alg:tdlambda}
   \begin{algorithmic}[1]
			\Function{TD($\lambda$)Update}{$s'_t, \delta_t$}  
			\If {($!T_{lrnRM}$ \& $s'_t$ generated by random action)} 
					\textbf{return}
			\EndIf
			\For {$k=t$ downto $\max(t-h,0)$} 		\Comment for all states in $s'_t$'s horizon
					\For {$i=1$ to $m$}			\Comment for all n-tuples
							\State $M \leftarrow \{\}$	
							\For {$s' \in S(s'_k)$}
									\If {$Ind_i(s') \notin M$}
										\State $w_i[Ind_i(s')]  \leftarrow w_i[Ind_i(s')] + \frac{1}{mN_S} \alpha\delta_t \lambda^{t-k} \sigma'\left(\nu(s'_k)\right)$
										\State $M \leftarrow M \cup \{Ind_i(s')\}$		\Comment inhibit index $Ind_i(s')$
									\EndIf
							\EndFor
					\EndFor
			\EndFor
			\EndFunction
	\end{algorithmic}
\end{algorithm}

\begin{table}%
\caption{Symbols of the algorithms and their source code equivalent in GBG}
\label{tab:symbSource}
\noindent
\begin{tabular}{l|p{8.5cm}}
  \hline
	symbol				& source code equivalent \\
	\hline
	\textsc{MakeAction} 				& constructor \texttt{NextState.NextState(state,action)} \\
	\textsc{ComputeAfterstate} 	& \texttt{StateObservation.advanceDeterministic()} \\
	\textsc{AddRandomPart}	 		& \texttt{StateObservation.advanceNondeterministic()} \\
	\textsc{TD($\lambda$)Update}& \texttt{NTuple2ValueFunc.update()} \\
	\textsc{TD-FARL-Episode}	& \texttt{TDNTuple3Agt.trainAgent()} \\
	TD-algo from~\cite{Bagh15,Kone15c}	& \texttt{TDNTuple2Agt.trainAgent()} (deprecated) \\
	\textsc{Sarsa-FARL-Episode}& \texttt{SarsaAgt.trainAgent()} \\
	$T_{after}$  & \texttt{AFTERSTATE}, \texttt{TDNTuple3Agt.getAFTERSTATE()} \\
	$T_{lrnRM}$  & \texttt{learnFromRM} in \texttt{TDNTuple3Agt.trainAgent()} 
\end{tabular}
\end{table}

\subsection{Temporal Coherence Learning (TCL)}
\label{sec:TCL}

The TCL algorithm developed by Beal and Smith~\cite{Beal99} is an extension of TD learning. It has an adjustable learning rate $\alpha_i$ for every weight $w_i$ and a global constant $\alpha$. TCL works by replacing the global learning rate $\alpha$ in Step 8 of Algorithm~\ref{alg:tdlambda} for each weight by the weight-individual product $\alpha\alpha_i$. The main idea behind TCL is pretty simple: For each weight, two counters $N_i$ and $A_i$ accumulate the sum of weight changes and sum of absolute weight changes. If all weight changes have the same sign, then $\alpha_i=|N_i|/A_i=1$ and the learning rate stays at its upper bound. If weight changes have alternating signs, then $\alpha_i=|N_i|/A_i \rightarrow 0$ for $t \rightarrow \infty$, and the learning rate will be largely reduced for this weight. 

The full TCL algorithm with all its variants is found in~\cite{Bagh15}. Of particular importance is the variant TCL-EXP that we newly developed in~\cite{Bagh15}: We introduced an exponential transfer function
	\begin{equation}
			g(x)=e^{\beta(x-1)}
	\label{eq:tclExp}
	\end{equation}
and set $\alpha_i=g(|N_i|/A_i)$. It was shown that TCL-EXP results on the game ConnectFour in much faster learning and higher win rates than plain TCL~\cite{Bagh15}.  

\subsection{Parameter Settings}
\label{sec:paramExp}

In this appendix we list all parameter settings for the experiments in Tab.~\ref{tab:allGames}. Parameters were manually tuned with two goals in mind: (a) to reach high-quality results and (b) to reach stable (robust) performance when conducting multiple training runs with different random seeds. The detailed meaning of all parameters is explained in the GBG documentation~\cite{Konen20a}. 

\subsubsection{2048:}
$\epsilon=0$, $\alpha=1.0$, $\lambda=0$, NO output sigmoid, USESYMMETRY, AFTERSTATE, NORMALIZE=false, FIXEDNTUPLEMODE=2 \cite[Fig. 3c, 4 6-tuple]{jaskowski2018mastering}, TC-id with TC-Init$=10^{-4}$ and  recommended weight-change accumulation. 200.000 training episodes, numEval=5.000.

\subsubsection{TicTacToe:}
$\epsilon=0.1 \rightarrow 0.0$, $\alpha=1.0 \rightarrow 0.5$, $\lambda=0$, output sigmoid, USE\-SYMMETRY, NORMALIZE=false, one random 9-tuple, no TCL. 30.000 training episodes, numEval=1.000.

\subsubsection{ConnectFour:}
$\epsilon=0.1 \rightarrow 0.0$, $\alpha=3.7$, $\lambda=0$, output sigmoid, USE\-SYMMETRY, NORMALIZE=false, fixed-n-tuple mode 1: 70 8-tuples. TC-EXP with TC-INIT$=10^{-4}$, $\beta=2.7$ and recommended weight-change accumulation. 5.000.000 training episodes, numEval=100.000.

\subsubsection{Hex 6x6:}
In our experiments, Hex is played on a 6x6 hexagonal board.

$\epsilon=0.2$, $\alpha=1.0 \rightarrow 0.5$, $\lambda=0$, output sigmoid, USESYMMETRY, NORMALIZE=false, 25 random 6-tuples, TC-id with TC-Init$=10^{-4}$ and  recommended weight-change accumulation. Choose-Start-01 and Learn-from-RM active. 300.000 training episodes, numEval=10.000.

\subsubsection{Othello 8x8:}
$\epsilon=0.2$, $\alpha=0.2$, $\lambda=0.5$, horizon cut 0.01, eligibility trace ET, output sigmoid, USESYMMETRY, fixed n-tuple mode 6: 50 6-tuples. TC-EXP with TC-INIT$=10^{-4}$, $\beta=2.7$ and recommended weight-change accumulation. Choose-Start-01 active. 250.000 training episodes, numEval=2.000.

Evaluation is done by playing episodes from all 244 4-ply start states in both roles against selected opponents. If both agents played perfect, the expected win rate would be 50\%. 

\subsubsection{Nim 3x5:}

$\epsilon=0.1$, $\alpha=0.5$, $\lambda=0.5$, horizon cut 0.1, eligibility trace ET, output sigmoid, n-tuple fixed mode 1: one 3-tuple. TC-id with TC-Init$=10^{-4}$ and  recommended weight-change accumulation. Choose-Start-01 inactive. Learn-from-RM active. 20.000 training episodes, numEval=1.000.

\subsubsection{Nim3P 3x5:}
In our experiments, Nim3P (3-player variant) is played with 3 heaps of size 5 (3x5) and the extra rule being active.

$\epsilon=0.15$, $\alpha=0.2$, $\lambda=0.5$, horizon cut 0.01, eligibility trace with RESET, output sigmoid, n-tuple fixed mode 2: two 3-tuples. TC-id with TC-Init$=10^{-4}$ and  recommended weight-change accumulation. Choose-Start-01 active. Learn-from-RM inactive. 300.000 training episodes, numEval=10.000.

\end{document}